# AOP-Wiki EMOD 3.0: Data Model Expansions and Content Evaluation Framework for Using Agentic AI to Improve Integration between AOPs and New Approach Methodologies (NAMs)

Virginia K. Hench [0000-0002-3612-8036][1], J. Harry Caufield [0000-0001-5705-7831][2], Sierra A.T. Moxon [0000-0002-8719-7760][2] , Jason M. O'Brien [0000-0001-8431-5824][3] , and Stephen W. Edwards [0000-0002-7985-111X][4]

1 Open BioData Modeling, Raleigh, NC, 27609, USA
2 Environmental Genomics and Systems Biology, Lawrence Berkeley National Laboratory, Berkeley, CA 94720, USA
3 National Wildlife Research Centre, Environment and Climate Change Canada, Ottawa K1S 5R2, Canada.
4 UL Research Institutes - Chemical Insights, Evanston, Illinois, 60201, USA

## Abstract

Adverse Outcome Pathways (AOP) are logic models that causally link biological mechanisms that can be measured in a lab to adverse outcomes, relevant to chemical regulatory endpoints. AOPs contextualize new approach methodologies (NAMs), in vitro and in silico methods used as alternatives to animal testing and the sequential events in an AOP serve as multi-scale models spanning biological scales. The AOP-Wiki serves as the global repository for AOPs. While the AOP-Wiki has played a central role in AOP expansion over the past decade, constraints within the current data model and application infrastructure limit the AOP-Wiki from supporting continued AOP growth and evolution. Yet, the transformative power of agentic AI has re-invigorated AOP-Wiki data modernization efforts at a time when core AOP principles can be harnessed to inform use of AI for aggregating and structuring AOP-relevant information. Seizing upon this momentum, we present AOP-Wiki EMOD 3.0, the third in a series of evidence model prototypes, which concretely demonstrates data model expansions and our vision for how the AOP-Wiki might be transformed to better serve regulatory science and emergent use of AOPs in biomedical and One Health contexts. We aim to lay a foundation to support computationally-generated AOPs and quantitative AOPs (qAOPs) by focussing on solutions for AOP-Wiki internal quality improvement, evidence structuring to enhance AOP FAIRness and AI-readiness, and improved integration between the AOP framework and NAMs to better serve next generation risk assessment.

## Introduction

The AOP-Wiki was first released publicly for beta testing in 2013 with the initial production release in 2014[1–4] as the global repository for Adverse Outcome Pathways (AOPs), a causal pathway framework introduced in 2010[5]. AOPs give context for new approach methods (NAMs) by serving as logic models for translating measurements from *in vitro* and *in silico* experimental methods to individual and population-level outcomes. AOP expansion within the AOP-Wiki and evolution of the AOP framework have been catalyzed by support from the Organization for Economic Cooperation and Development (OECD)[6,7] and the Society for Advancement of AOPs (SAAOP)[8].

AOPs are used to model causal biological mechanisms for a broad range of chemical safety uses in toxicology, as well as non-chemical hazard and risk assessment and support regulatory needs by providing foundational mechanistic toxicology knowledge for Integrated Approaches to Testing and Assessment (IATA)[9,10]. For example, the first OECD-endorsed AOP has fostered development of numerous *in chemico* and *in vitro* assays and test guidelines used for skin sensitization hazard and risk assessment[11]. More recently, OECD endorsed a set of AOPs triggered by radiation exposure events leading to adverse outcomes (AOs) covering vascular remodeling, cataracts, bone loss, and learning and memory[12,13].

Importantly, use of the AOP framework outside of the OECD AOP Programme and outside the AOP-Wiki invites questions about the role of the AOP-Wiki and the value of the AOP framework as a conceptual tool for evidence assessment and causal modeling. On a more practical level, the existence of AOPs reported in the scientific literature, but not incorporated into the AOP-Wiki[14], influence our motivation to modernize the AOP-Wiki with EMOD 3.0 to reduce barriers to AOP entry and to align with data-driven approaches to AOP development. We intend the AOP-Wiki to be a readily usable data resource that benefits from agentic AI methods[15]. We present an AI-based approach for reducing redundancies in the AOP-Wiki that interfere with AOP usefulness. The inclusion of rigorously assessed supporting evidence, trustworthy causal models, and detailed provenance all support the goal of enriching AOP-Wiki's compatibility with AI methods while maintaining the central role of domain experts in curating and validating its contents.

The AOP-Wiki was originally built as part of an international collaboration to create a central AOP knowledgebase (AOP-KB) to support use of NAMs and NAM-based decisions around the world. While many other tools exist for using the data from the AOP-KB, the AOP-Wiki is currently the only authoring tool for submitting new content. It was originally developed at the US Environmental Protection Agency (EPA) between 2012-2018. Since 2020, maintenance and continued evolution of the AOP-Wiki has been driven by AOP community working groups and supported by the European Commission's Joint Research Centre (JRC), yielding important additions like content licensing updates, incorporation of the OECD's AOP Developer's Handbook, exposure of bio-ontology details, pagination to reduce slow page load times, and more[16,17]. Due to the number of organizations involved in the collaborative AOP knowledgebase project, the SAAOP was formed to host the AOP-Wiki and promote its use for collaborative AOP development[1].

Yet, requests for user-facing enhancements like AOP visualization standards and more intuitive interfaces for tracking contributor attribution and references around 2021-2022 prompted interrogation of the limits of the current AOP-Wiki infrastructure. While some aspects of the AOP-Wiki's Ruby/Rails codebase offer a strong foundation for iterative changes to the MySQL-based database model, the close coupling between backend and frontend components limited rapid prototyping of the user-facing features that have been requested. Around the same time period, a desire to improve handling of AOP-supporting evidence within the AOP-Wiki yielded the first AOP-Wiki Evidence Model (EMOD) prototype, with funding support from Climate Change Health Canada, the Alternatives Research & Development Foundation, and the Canadian Centre for Alternatives to Animal Methods. EMOD version 1.0 and a subsequent pair of prototypes expanded the current AOP-Wiki code base and data model and allowed for useful insights to be collected, but the need for a more modernized system remained[18,19].

To address limitations to the current AOP-Wiki, this report introduces AOP-Wiki EMOD 3.0, a completely new AOP-Wiki prototype web application that persists and expands core elements of the data model using separate backend and frontend web application frameworks. By isolating the frontend and backend components, AOP-Wiki EMOD 3.0 provides a modular foundation for future revisions and expansions that will eventually replace the current AOP-Wiki. We use EMOD to refer to a series of AOP-Wiki data model expansions and EMOD 3.0 to refer to the web application that has been released in 2026 to www.emod.aopwiki.org. EMOD 3.0 user interfaces (UI) are designed to showcase data model expansions and opportunities for content quality assessment and improvement. Through a series of selected domain-based use cases, we explain how EMOD 3.0 design decisions were made to address challenges and requests that have been raised in the AOP community over the past few years[8,20–27].

## AOP-Wiki Evolution

EMOD 3.0 continues an evolutionary trajectory of the AOP-Wiki from predominantly unstructured free-text entries to maximally structured content. Expanded entry templates for AOP-supporting evidence and methods in EMOD 3.0 will support annotation with a much broader variety of ontology terms to promote computable development approaches and data interoperability (Figure 1, panels B and C). While AOP-Wiki data model classes align with the core elements of the AOP framework — molecular initiating events (MIEs), key events (KEs), adverse outcomes (AOs), key event relationships (KERs), and AOPs (shown in Figure 1A), most properties on these data classes are free-text fields. The AOP-Wiki Release 2.2 added more structured elements to the AOP-Wiki data model with the introduction of Key Event Components (KECs)[28,29]. The KEC concept introduced use of OBO Foundry bio-ontology terms to AOPs, as a way of making KEs more computable. Each KEC defines a discrete action, object and process entity term. KEC Action terms are defined by an AOP-Wiki-specific controlled vocabulary that is based on the Phenotypes and Traits Ontology (PATO)[28]. The KEC biological object and biological process terms use selected bio-ontologies. Controlled vocabulary lists are used to define levels of biological organization (LoBO), sex terms, and life stage terms. CL terms and Uberon terms are used to define biological spatial context for KEs, and NCBI Taxon terms to label species applicability for KEs, KERs, and AOPs.Yet, while the bio-ontology-aligned expansions helped define KEs ontologically, evidence for causality, represented by KERs that link upstream KEs to downstream KEs, has not been structured in a manner that enables scalable tracking of KER evidence provenance, an issue that the first EMOD prototype aimed to address[17]. EMOD v1.0, introduced a new set of data classes for structuring lines of evidence supporting KERs. The Evidence and Observation classes were established to structure AOP-supporting evidence from peer-reviewed studies using the same bio-ontology term lists used for the KECs, LoBO, and the sex, life stage and taxonomic domains of applicability (DoA), represented by the gold boxes in Figure 1B and in Supplemental Figure 1A.

EMOD v1.0 provided a foundation for data model concepts embraced by the international Methods2AOP collaborative, a group that formed to identify ways of improving integration between AOPs and NAMs[26,27] by focussing on structuring content associated with the KE free text property, “How is it Measured or Detected”, represented in Supplemental Figure 1B, a precursor figure that predated the concept model included in the Methods2AOP collaborative publication[27]. The M2AOP prototype introduced an Assay data class that has properties essential for structuring the biological elements measured by a given assay or NAM (Supplemental Figure 1C). To promote AOP-Wiki internal mapping between Assays and KEs based on the biological components measured by NAMs, the Assay model includes biological object and process properties, populated using the same bio-ontology term lists available for the KECs (Figure 1C).

Data model-aligned webforms in the M2AOP prototype enabled stress testing by Methods2AOP members. Stress tester feedback highlighted ways that some M2AOP data classes deviated from the central purpose of the AOP-Wiki to serve as a knowledge base for AOPs. For example, contributions to the Study Design data object had long, detailed free text descriptions akin to details found in methodology sections of peer-reviewed publications and test guideline reports. This kind of information does not need to be detailed within the AOP-Wiki, but should be referenced via linkouts to referenced sources as a form of provenance tracking. Stress testing exposed a need for clarifying the meaning of “methods” to avoid conflation of assays with test guidelines. An EMOD 2.0 prototype was built to integrate EMOD v1.0 with insights collected from the M2AOP v1.0 stress testing and was publicly deployed for approximately two years but was never tested by users[19]. Supplemental Figure 1D provides a graphical summary of the EMODv2.0 prototype data model.

## AOP-Wiki EMOD 3.0

AOP-Wiki EMOD 3.0 offers a new web application that expands and refines prior EMOD prototype data classes, incorporates use case content to demonstrate the value of the new data classes, and adds approaches for promoting content quality assessment, paving new directions for AOP-Wiki content improvement and internal biological consistency monitoring using scalable and computable approaches. Use cases have been selected to showcase a diversity of ways that the scientific community is using the AOP framework to support advancement of NAMs and next generation risk assessment (NGRA). Incorporation of completion metrics for AOPs, KEs, and KERs, as well as a new Event Integration Score, provide quantifiable benchmarks to inform priorities for annotation and more in-depth evidence analysis. Data models and page layouts developed to hold large language model (LLM)-based KE groupings help cluster potentially redundant KEs within the AOP-Wiki, a well-known pain point in the AOP community. EMOD 3.0 showcases KEs and AOPs that were harmonized for a NICEATM-led study focussed on identifying seizure targets for chemical and drug safety testing[30]. The candidate KE merger groups derived from LLM-based analysis of KE titles and from a lung fibrosis genomics study[31] are presented in a manner to catalyze needed discussions about AOP standards and evolution of AOP ownership over time. Lastly, EMOD 3.0 includes proof-of-concept examples of contributor tagging and provenance tracking, two features collected by SAAOP as needed AOP-Wiki 3.0 feature requirements and presented to the OECD EAGMST[8]. Alongside EMOD 3.0, we have released an AOP-Wiki CLI application that includes scripts for processing the AOP-Wiki XML, enriching for preferred subsets of AOP-Wiki entities, and exporting the processed entities as JSON, CSV, and Excel formats, all including the entity completion scores and Event integration score.

### *EMOD 3.0 Use Cases Covering Integration between AOPs and NAMs*

Selection of the use cases featured in EMOD 3.0 was informed by our experience working in the AOP community, by scientific literature covering AOPs and NAMs, and discussions with biomedical researchers newly interested in AOPs due to their association with NAMs. We acknowledge demands for the AOP-Wiki to better support how AOPs are used in regulatory contexts, but prioritized the needs of potential AOP-Wiki contributors and barriers to expanding and improving AOP-Wiki content. We are also aware of the growing number of tools that restructure and reformat AOPs from the AOP-Wiki to better serve chemical safety and food safety regulator's needs[32–35]. Yet, amidst AOP Knowledge Base (AOP-KB) associated tools, the AOP-Wiki is the sole tool for authoring AOPs and for this reason, it has an important role to contribute with respect to integrating insights from the AOP and NAMs communities and for upholding AOP quality standards. By presenting custom data model classes and page views for the biomedical use cases listed below, EMOD 3.0 offers ways of framing challenges facing the AOP community in a manner that will lead to solutions that enhance the AOP-Wiki's position as a resource for advancing NAMs.

#### *Depression and Neural Network Use Case*

As part of a long-range vision for how AOPs can inform strategies for transitioning away from *in vivo* models to NAMs, we explored the Depression and Neural Network Use Case. The use case was initiated in fall of 2025 during discussions on how Professor Rainbo Hultman and colleagues were using a mouse model of stress vulnerability and depression[36] to investigate pesticide chemical exposure impacts. The molecular and cellular mechanisms characterized in their *in vivo* toxicity model can be incorporated into an AOP and some mechanisms may already be represented in the AOP-Wiki. As science progresses, we can move in a direction whereby mechanistic understandings gained from *in vivo* studies form the basis for *in vitro* and *in silico* NAMs for chemical and drug safety screenings - a trajectory that the AOP framework can facilitate. To support this line of thinking, we searched AOP-Wiki v2.7 for AOPs and Events relevant to ongoing studies carried out by Hultman and colleagues. The challenges we encountered are representative of pain points that have circulated in the AOP-Wiki user community for many years (see Clerbeaux in 2023-2024 SKIG Report[24]). Specifically, this use case highlighted AOP-Wiki challenges that can be classified as KEs that are conceptual duplicates and KE's with titles suggesting they could be conceptual duplicates. At the AOP level, there are many AOPs that are sparsely populated without clear indication of their maturity beyond the OECD status assignment, which only pertains to a subset of all AOPs in the AOP-Wiki (highlighted in a recent SAAOP presentation[37]). A desire to reduce pain points impacting the whole AOP-Wiki user community and identify AOPs and KEs covering neuronal networks and mechanisms leading to depression motivated creation of the Event Integration Score (EIS), an Event to KER search interface, and an interface that allows queries for AOPs based on KE co-occurrence, without the KEs being paired in KERs.

*Seizure Use Case*

The seizure use case leveraged published curations from a seizure targets study led by the NTP Interagency Center for the Evaluation of Alternative Toxicological Methods (NICEATM) that aimed to define a set of biological targets for use in development of NAMs for drug and chemical safety testing. To support outputs from this study, we introduced new data classes for Harmonized Events, Harmonized AOPs, and Biological Target Families. Associations between chemicals and drugs known to induce or alleviate seizure were mapped to seizure KEs using the Observation data class.

*Lung Fibrosis Use Case*

The EMOD 3.0 lung fibrosis use case was motivated to align with the OECD Omics2AOP project, which aims to improve how AOPs are used to model omics data sets. The lung fibrosis use case page features groups of KEs clustered together based on a published gene expression analysis[31]. EMOD 3.0 displays the candidate KE merge groups, along with the associated AOPs that would be impacted by merging the KEs from each group.

## Methods

### Query and Content Review for the Depression and Neural Network Use Case

To support the Depression and Neural Network Use Case, queries were performed in the production AOP-Wiki v2.7 and using the AOP-Wiki database. A search through existing "depression" events revealed three Events, one of which stood out for having three associated AOPs, *Event 1346: Increased, depression*. At the time (December 2025), one of the Events was an orphan, meaning it was not associated with an AOP and a recently created event, *Event 2392: Depression*, was associated with a newer AOP, *AOP 604: Binding of Alpha 1-Adrenergics to Antagonists Leading to Depression*. Given the conceptual similarity between Events 1346 and 2392 and the AOP-Wiki's goal of promoting Event re-use across AOPs to form AOP networks, the AOP-Wiki Gardener's policy would be to alert the authors of AOP 604 to consider using Event 1346. Yet, closer inspection of the AOPs associated with Event 1346 revealed that all three were open for adoption , meaning the original AOP authors were no longer taking responsibility to continue maintaining the AOPs that were first created around the older Depression Event. Further, all three of the Open for Adoption AOPs that include Event 1346 were sparsely populated without any supporting references cited.

Aligned with the motivation of this use case to identify Events in the AOP-Wiki that align with Hultman's model that distinguishes stress vulnerability networks from depression pathology networks[36], we ran queries to identify Events describing neuronal networks and identified two, *386: Decrease of neuronal network function* and *618: Decreased, Neuronal network function in adult brain.* Content annotation needs associated with these two Events are covered in the Discussion section.

### EMOD 3.0 Web App Prototype Development

Development of the AOP-Wiki EMOD 3.0 Python FAST API app was initiated after using sqlacodegen[38] to generate Python SQLAlchemy models based on the EMOD v2.0 MySQL database. Data classes associated with administrative functions in the production AOP-Wiki code base were dropped. Later, to avoid bottlenecks associated with primary key and foreign key incompatibilities when running data table migrations, a content migration process that relies on JSON-formatted datadumps from the production AOP-Wiki was implemented. Uploader scripts within the EMOD 3.0 application were developed to import structured and annotated use case content. The frontend application is built with React. Claude Code and GitHub CoPilot using Claude models were used to write code and support code review.

### Use Case and AOP-Wiki Content Processing

The AOP-Wiki CLI App (CLI App) supports EMOD 3.0 by providing a set of text extraction functions that make use of the XML download available from the AOP-Wiki production site. Documentation on how to run the released code is available at https://github.com/gingin77/aop_wiki_cli; the functions most relevant to EMOD 3.0 are briefly described below. The README provides instructions on how to run the CLI functions in a development environment that has uv and Python installed.

#### *Entity Completion Scores and Event Integration Scores*

All CLI App functions start with making calls to fetch the AOP-Wiki XML and extract core entities - KEs, KERs, and AOPs - into Python dictionaries available as JSON files. Entity document completion scores are based on a simple percentage calculation based on the count of non-empty properties over the total number of document properties. The Event Integration Score offers a holistic approach for ranking Key Events by weighting and integrating multiple qualitative factors into a single numerical score. Details of the score calculation are available in the CLI App repo in the meta_data_helpers.py script. (https://github.com/gingin77/aop_wiki_cli/blob/5dc9aee0469bd6114036c0c35444fbfe8f40fe3b/src/analysis/meta_data_helpers.py#L27). Briefly, Event Integration Scores are decreased when the only associated AOPs have the Open for Adoption license and increased for the following factors: AOP Count, % Complete, OECD Endorsement, and being in the OECD AOP Programme. The Event Integration Scores are available through the JSON and CSV exports made available through the CLI **collect-event-integration-rankings** function and are also presented throughout the EMOD 3.0 web app. Importantly, the AOP-Wiki XML does not include orphan Events and KERs that do not belong to an AOP, whereas EMOD 3.0 includes all Events and KERs that have been created, similar to the production AOP-Wiki. Therefore, average completion scores for Events and KERs in EMOD 3.0 will not match average completion scores derived from the AOP-Wiki XML/CLI App.

#### *Assessment of Submitted KER Evidence Content*

Two CLI App functions offer ways to evaluate KER-supporting evidence submitted by AOP-Wiki contributors. All KER-specific CLI functions used KER extracted from the AOP-Wiki XML, which applies an HTML table element extraction functions to the main KER evidence fields: weight of evidence, empirical support, biological plausibility, and quantitative understanding. The **harmonize-ker-evidence** function takes all KERs and enriches for KERs that have tabulated evidence in one of those fields. A table column header harmonizer function then applies a standard for structuring KER-supporting evidence, in a manner that aligns with the EMOD Evidence data class. KER tabulated evidence results are then exported to an Excel workbook holding summary information and details from each KER with tabulated, harmonized evidence content. The **search-kers-for-concordance- text** CLI function provides a way to isolate KERs that have specific concordance evidence types mentioned in the evidence fields. This covers all mentions of incidence, temporal, and dose concordance in the free-text KER evidence fields. Results from the KER concordance search are exported to JSON and CSV files.

### *Seizure Use Case Extraction*

A single CLI command, **collect-harmonized-seizure-aops** orchestrates an extraction and comparison pipeline that transforms worksheets from the supplemental data published by Behl, *et al*[30] into mappings that were uploaded into EMOD 3.0. The processing pipeline details are documented in the CLI repo and the table below outlines the input worksheets used to generate mappings displayed on specific pages in EMOD 3.0. Mappings between Biological Target Families, ToxCast Assays and KEs were expanded by running a script to compare KE descriptions and Biological Targets from Behl against KE titles in the AOP-Wiki XML. The fuzzy-matching based approach was reviewed manually to accept or reject expanded mappings.

| **Supplemental Data Worksheet** | **Mappings obtained and page view in EMOD 3.0** |
| --- | --- |
| Suppl2_KEs AOP Harmonization | Mappings used to populate the harmonized AOP and harmonized KE pages |
| Suppl4_Compiled Compounds | Chemicals by CASRN with direction of effect, PubChem evidence flag, and parsed literature citations |
| Suppl6_ICE Assays | Contains mappings displayed on the Biological Target Families page, which includes mappings between Biological Target Families, ToxCast Assays and KEs. |

### *Lung Fibrosis Use Case Content Preparation*

The Event groupings for the lung fibrosis use case were collected from Figure 4 in Saarimäki, et al[31] and organized into a JSON file, allowing ingestion into the Candidate Event Merger data class, using one of the data uploader scripts in the EMOD 3.0 code base.

## Results

The EMOD data model elements will be described first, followed by EMOD 3.0 UI features that have been introduced to promote AOP-Wiki content quality assessment and improvement. The content structuring allowed by the data model expansions and the content quality improvement tools are both needed to best serve the AOP-Wiki user community and fully harness the power of agentic AI in support of AOP development and expanded use of AOPs in support of NAMs and NGRA.

### AOP-Wiki EMOD 3.0 Data Model Expansions and Principles.

Three key principles introduced to the AOP-Wiki by the EMOD model are illustrated in Figure 2 and will be detailed below. Upward pointing arrows represent the roll up principles and the intended flow of information from the Observation and Evidence data classes. The Observation class is directly associated with a KE and an Evidence data class that holds two Observation data objects and is directly associated with a KER.

### *Roll-up Principle*

The roll-up principle refers to how properties on child data objects are accessible on parent data objects, such that, in the context of the AOP-Wiki, observation and evidence information submitted to KEs and KERs should be available to parent AOPs and contribute to defining the AOP's properties. The roll-up principle aligns with the method chaining concept used in object-oriented programming and if fully implemented would support principles developed for making taxonomic domain of applicability assignments for AOPs[39]. Although we do not yet have a fully populated example to demonstrate the roll-up principle, it is an important concept influencing the data classes in EMOD.

### *Causal Agent*

The causal agent concept helps distinguish biological elements applied in experimental systems to induce perturbations from biological events that occur in the closed biological systems that AOPs represent.

Causal agents include chemical and non-chemical stressors (like viral pathogens or radiation), along with biological entities that are applied in experimental systems. The causal agent concept was developed in the context of an AOP-based text annotation effort where there was a need to differentiate biological entities applied in experimental systems from biological entities that could be biological objects represented as AOP KEs[19]. For example, to understand the effects of a cytokine in the immune system, a researcher might add the cytokine to an experimental system where it acts as a causal factor triggering measurable biological processes like gene transcription or cell proliferation. This type of experimental setup helps define the causal relationship between biological entities and processes, whereby the biological entity that functions as a causal agent in the experimental system represents a biological object in an upstream KE, which influences another biological process represented by a downstream KE.

### *Observation*

The purpose of the Observation data object is to inventory and structure empirical evidence for the KE. Figure 3 shows how the EMOD 3.0 Observations table holds associations between chemical stressors and seizure occurrence. Use of the Experimental Effect and Phenotype properties together differentiate chemicals with seizure therapeutic potential from seizure-inducing chemicals, which informs whether the Observation maps to the KE for Decreased Seizure vs Seizure.

More generally, AOP Observations model causal associations between stressors and KEs and may include an explicit structured method of measurement/associated Assay data object. Figure 2 lists Observation data class properties for defining a biological object, process and/or phenotype to serve KEs covering all levels of biological organization — from molecular level to population level. Structuring KE-associated Observations enables semantic annotation with bio-ontology terms, which can be used to build AOP-Wiki internal biological consistency checks and interoperability with other biomedical knowledge resources, which is represented by the more detailed and ontological representation of both Observation and Assay data object properties Figure 3A. The Observation's experimental effect property is set up to be populated with the same controlled vocabulary list that is used for KEC Action terms.

### *Assay*

The Assay data class in the EMOD 3.0 data model builds upon earlier prototypes and recommendations from the Methods2AOP collaborative[26,27] that have been discussed above and are graphically represented in Supplement Figure 1. Yet, the latest EMOD data model is built to better accommodate complex *in vitro* systems, which model endpoints representing more than one KE in an AOP. For example, a study from Schmidt and Suter-Dick[40] used a liver fibrosis microphysiological system to model multiple KEs from the liver fibrosis AOP #38, which is shown in Figure 4. The gene expression measurements align with KEs and could be represented as Observation data objects that all use the same liver microphysiological system as a NAM. The biological properties of the Assay data class – measured biological object, process and phenotype – are depicted in Figure 1C.

### *Evidence*

The Evidence data object is designed to structure causal evidence for a KER by linking together two Observation data objects that are aligned with the KER's upstream and downstream KE's[18], as is illustrated in Figure 2. When the Evidence data class was introduced as part of the first EMOD prototype, its properties were selected to align with an Evidence Map interface. The Evidence Map feature was designed to present a visual summary of KER evidence, broken down by biological plausibility and concordance type — temporal, dose and incidence. The Evidence Map had a row for each submitted literature reference and a way of rating whether the reference provided supporting or conflicting evidence based on the evidence type column. A snapshot of the EMOD v1.0 Evidence Map feature, populated with mock data, was presented in a poster at the 2023 Society of Toxicology meeting[18] and is available as Supplementary Figure 2. The Evidence data class was not part of the M2AOP prototype that was stress tested by multiple members of the Methods2AOP collaboration, so the Evidence class has not undergone stress testing beyond a much smaller group of AOP developers.

Rather than focus on creating a new Evidence Map or other Evidence-focussed user interface feature in EMOD 3.0, attention has been devoted to analysing the content submitted to AOP-Wiki KER evidence fields, by creating the KER evidence analysis scripts in the AOP-Wiki CLI app and reviewing the outputs. The harmonize-ker- evidence function captured KERs with HTML table elements in the evidence fields. Importantly, some KER entries with HTML table elements are based on the recommendations in the AOP Developer's Handbook[41], but the KER contributors are free to define the exact elements used in the HTML tables that they create so further harmonization is needed to transform submitted KER evidence content to align with the template provided by the KER Evidence class. The harmonization process implemented a series of input table header comparisons against a pre-defined set of harmonized headers. Out of 2336 KERs, 183 had tabulated entries in the evidence fields and 52 of the entries were harmonizable using the scripts called by the harmonize-ker-evidence AOP-Wiki CLI function. We also searched KER entries for explicit mentions of temporal, dose or incidence concordance in the evidence fields. Only 139 KERs had matches and the snippets holding the matches tended to not offer the type of biological contextual detail that would fit into the Evidence class. The exports from the AOP-Wiki CLI KER functions are available as supplemental data files.

From this analysis of the submitted KER free text evidence content, we concluded that more use cases involving the subject matter experts who have contributed to KERs and AOPs are needed to support refinement of the Evidence class properties. Meanwhile, an Evidence class is featured in the AOP EMOD LinkML schema that is still undergoing development[42], alongside EMOD 3.0.

### *Experiment Type*

The Experiment Type data class serves as an AOP-Wiki internal controlled vocabulary for defining the type of study performed. The list currently includes the following terms: *in vivo, in vitro, in silico, ex vivo, in situ,* cell-free, and biochemical. Terms relevant to population-level analysis like epidemiological study or clinical study have been discussed, but not yet incorporated.

### *Citation*

The EMOD Citation data object provides a structure for all References free text fields on AOP, KER and KE data objects and is critical for enhancing provenance tracking within the AOP-Wiki. Citation properties are the fields typically needed for journal citations and published books, including URL link fields for DOI's and PubMed IDs.

### *Biological Target Family*

A Biological Target Family class was introduced as part of the seizure use case (see Supplemental Figure 3), so the current list of targets is scoped based on association with seizure. Biological targets tend to align with the biological objects in MIEs. Therefore, by incorporating a list of biological targets directly into the AOP-Wiki, EMOD 3.0 enables AOP-Wiki internal assessment of AOP coverage based on MIEs.

*Table 1: AOP EMOD Data Classes* Summary Table*

| EMOD Data Class | Role Compared to AOP-Wiki v2.8 |
|---|---|
| Observation | New entity used to structure the empirical evidence for associations between stressors and KEs, which are sometimes included in the KE descriptions, as well as in statements that provide justification for creating an AOP. |
| Assay | New entity that replaces the KE free text field, *How is it measured or detected,* and can be directly related to AOPs, KERs, and Observations |
| Evidence | New entity that structures and supplements free text KER fields: *Empirical Support, Quantitative Understanding, Weight of Evidence,* and *Biological Plausibility* |

| Experiment Type | New entity that serves as a pick-list for defining Assay, Observation and Evidence data objects. |
|---|---|
| Citation | New entity that structures individual references in place of the free text Reference field on KEs, KERs, and AOPs |
| Biological Target Family | A new entity added for the purpose of AOP-Wiki internal AOP coverage evaluation against a fixed list of biological targets. |
| ** EMOD 3.0 includes other new data classes for bulk import, AOP harmonization, and grouping of KEs, but these are considered meta classes needed by the AOP-Wiki that supplement AOPs, so are not included here.* | |

## EMOD 3.0 Content Quality Assessment and Improvement Features

### *EMOD 3.0 Landing Page and Document Completion Scores*

The EMOD 3.0 landing page, shown in Figure 5, features three main page types — Core Entities, Use Cases, and Advanced Search — along with a panel for Average Document Completion Scores. Each KE, KER and AOP link leads to a search interface that offers filter and sort options. The LLM-driven Event Group pages provide examples of seemingly redundant Events in the Merger Groups and KE pairs that exemplify the Same Object, Different Action, or SODA principle developed in one of the 2018 AOP Network papers[43]. The Completion Score panel shows how Events have the highest average completion scores, followed by AOPs, and then KERs. Harmonized Events and AOPs that were added as part of the Seizure Use Case do not have content, so they lower the average completion scores.

### *Event Integration Score (EIS) and Sorting Features*

The Event Integration Score (EIS) offers a single quantitative measure that incorporates multiple factors associated with seemingly redundant Events. The EIS increases based on the Event's associated AOP count, whether the associated AOP's are endorsed by OECD, the Event Completion Score, and whether there is content submitted in the Event's *How is it Measured or Detected* field. The EIS is reduced when all associated AOPs are Open for Adoption.

EIS scores are represented throughout EMOD 3.0 Event pages as the "Integration Score" as well as on the LLM-Driven Event Group pages.

The EMOD 3.0 Events Search page includes multiple options for searching, filtering and sorting Events, along with the ability to dynamically add columns in support of more complex Event sorting and comparisons. The Event search page snapshot in Figure 6 shows the full set of columns available and default column and sort settings with a sort on the Event ID column. A focus on the four Events shown in Figure 6 demonstrates the purpose of the EIS with *Event 3: Reduction, 17beta-estradiol synthesis by ovarian granulosa cells* having the highest EIS out of the four Events shown. Event 3 has the highest EIS because it is associated with nine AOPs, two of which are endorsed by the OECD, has a high completion score, and has a completed entry describing how it is measured, indicated by the check mark in the "Has Method Text" column. In contrast, *Event 9, Activation, 5HT2c* has a negative EIS because it has a low completion score, doesn't have any text describing how it is measured, and is only associated with one AOP, which is Open for Adoption. The Open for Adoption detail for the AOP associated with Event 9 is not shown in the figure, but could be revealed in the interface by clicking the toggle option or by checking the box for "# AOPs Open for Adoption" to add that column. The purpose of exposing this kind of Event metadata in a searchable form is to promote strategic approaches to managing and improving the content held in the AOP-Wiki.

### *KE Groupings to Support AOP-Wiki Content Clean-Up*

In the Intro section on the Depression and Neural Network use case, we noted a problem with redundant KEs and how the issue of KE redundancies has been raised many times in the AOP-Wiki user community. Another issue is that some KEs that have titles that suggest redundancy but the details of their content indicate nuanced differences not reflected in the title. The issue of poor KE titles can be addressed with annotation efforts aimed at applying bio-ontology terms based on the free text content to differentiate KEs based on cell or organ location or life stage, sex or species domains of applicability. However, resolution of cases where KEs are sufficiently similar to merge would reduce the number of KEs to annotate and save on compute resources.

We have used EMOD 3.0 to showcase four distinct approaches for KE groupings (Figure 7). Two approaches from the seizure[30] and lung fibrosis[31] use cases involve KE groupings reported in the scientific literature as part of efforts to use AOPs to support NAMs development. To reach their goal of obtaining a discrete set of seizure pathway targets to use in support of identifying NAMs, the seizure targets study authors needed to reduce the total number of KE associated with the seizure AOPs that they identified in the AOP-Wiki. To resolve their issue, they manually curated source KEs to merge into newly named harmonized KEs[30] (Figure 7A). In the lung fibrosis NAMs study from Saarimäki, *et al.,* a genomics approach was used to identify similar KEs based on gene expression signatures, but a preferred KE wasn't suggested for harmonization. Rather than force harmonization without a named KE for harmonization, these KEs are treated as candidates for merging ((Figure 7B). The KE grouping approaches performed by domain experts for specific contexts of use are valuable insights that can benefit the wider AOP-Wiki user community. However, a more scalable approach is needed to address KE redundancies across the entire AOP-Wiki, so we used an LLM based approach to identify potential KE duplicates and KEs known as SODA KEs, which represent the <u>s</u>ame biological <u>o</u>bject, with a <u>di</u>fferent <u>a</u>ction. The LLM-based groupings are represented in Figure 7, panels C and D. In each the EIS is shown at the end of each KE title. Panel C includes an instance where a preferred KE has been manually curated and is labelled. The individual KE grouping pages provide the rationales for each group and the rationale for a preferred KE selection.

## Discussion

The AOP-Wiki EMOD 3.0 core data model expansions and infrastructure improvements are intended as a first step towards structuring AOP-Wiki content as part of a larger data modernization effort, informed by many community engagement efforts[21,22,44–46] and stakeholders[8,24]. Further, we hope that these technical transformations will help advance AOP Standards, which are expected to involve more use cases and input from the wider AOP and NAMs communities. The EIS and AOP document completeness scores that are presented in EMOD 3.0 provide a semi-quantitative framework to support AOP-Wiki content organization and prioritization for agentic annotation approaches, but application of this framework towards quality improvement will require community engagement and support from other AOP-Wiki stakeholders. Similarly, any KE harmonization and execution of KE de-duplication efforts suggested by the candidate KE merger groupings, will require collaboration with the AOP-Wiki Gardening team and coordination with AOP authors.

### EMOD 3.0 and Future Directions for the AOP-Wiki

For almost 15 years, the AOP-Wiki has fulfilled its critical role as the internationally recognized repository for AOP information. The information contained in the underlying AOP-KB is being incorporated into an ever-growing number of third party tools[33,35,47,48] including tools that directly support risk assessment decisions. However, the wiki-inspired format for capturing and storing this information is no longer sufficient to support modern uses of this information. The EMOD project goes far beyond a simple software upgrade and reimagines the AOP-Wiki (and by extension AOP-KB) data model to meet the current needs. The inspiration for these changes comes from user community feedback as well as the current best practices for knowledgebase development.

Consistent with the directions provided in the OECD supported AOP Developer's Handbook, AOP developers typically have a specific context of use driving the decisions made when defining an AOP. At the same time, many different developers will describe the same biological processes within slightly different contexts. This creates a tension between the desire to have a single description of the biological system and the different perturbations of that system and desire of an AOP developer to have a narrative that is narrowly focused on the given context. The original wiki format makes it difficult to alleviate this tension because changes to a KE or KER description to support one context may be inconsistent with the original context. As a result, separate KEs (and KERs) are defined that describe the same fundamental biology in slightly different ways. By breaking these descriptions into discrete entities with the proper semantics, we can explicitly capture the similarities and differences between KEs thereby allowing individual AOP developers to precisely define the KEs within the context of their AOP while still allowing systems models to merge those semantically similar KEs when traversing an AOP network to understand the core biological concepts.

The way of managing authorship within the wiki framework causes problems as well. In a traditional wiki, all content is open for editing and authorship is tracked through the revision history of the individual pages. From the beginning, the AOP-Wiki has restricted this on AOP pages due to the context issues discussed above. With more emphasis on integrated systems modeling, this practice is not sustainable. It also eliminates the option of having contributions coming from AOP authoring tools because of change conflicts arising from edits to the same block of text.

The new EMOD 3.0 infrastructure begins to address the challenges associated with collaborative content authoring. The data model elements that support the provenance tracking features associated with bulk upload of the seizure use case observations serve as a foundation for breaking individual author contributions into discrete entities, which will make authorship more readily apparent and allow for many tools to contribute content in parallel. In addition, the EMOD evidence layer – specifically, the Observation, Assay, and KER Evidence classes – enable information capture at the different levels to make KE and KER descriptions less dependent on the context of a specific AOP.

The EMOD Observation and Assay classes allow more explicit description of KEs based on the methods used to measure the event, the biological objects and processes that are perturbed, and the observations of those perturbations that have been made using the methods, which is a key improvement pertaining to integration between AOPs and NAMs. KERs then focus on the paired observations across two KEs and an evaluation of the collective evidence from these observations, coupled with biological plausibility given the biological objects and processes involved, that a causal relationship exists. Following the roll-up principle, an AOP will then be defined through these objective descriptions and provide the proper context for a specific decision. For integrated systems models, the KEs and KERs can be used independent of the AOP context to better understand the overall biological system under different perturbation scenarios. By providing another layer for modeling AOP-supporting evidence, we are providing the infrastructure needed to realize a vision for AOPs as multiscale models that has been developing over many years[49], which facilitates tracking of emergent evidence for defining novel AOPs.

By explicitly defining methods and the observations using those methods, the EMOD data model provides a robust framework for organizing NAMs information. Given the central role of the AOP framework in development of IATA to support regulatory decisions, having this information captured in a programmatically accessible format opens the door for advanced computational models to support regulatory decisions. Beyond that, the EMOD model makes the information much more searchable for individual users simply trying to understand the potential uses for a given method.

While AOP-Wiki development has always been guided by the FAIR principles, the new data model allows for better semantic alignment, enabling computable approaches for tracking biological consistency between components. For findability, the increased use of ontological terms in a carefully designed semantic framework increases the chances that a user will find the needed information regardless of the diversity of terminology across disciplines. Interoperability is the biggest change with the new data model. All entities will be described using external, widely adopted ontologies with minimal free text descriptions. Since data from the AOP-Wiki have been programmatically available for a decade via XML formats, this transformation in how the information within the wiki is organized will allow outside tools and models to directly consume structured AOP information and combine it with any other data annotated with compatible ontologies. For reusability, KE and KER content have always been under the BY-SA share license, allowing reuse as soon as the data were entered into the AOP-Wiki. The new structure sets a foundation for information entered on one entity, such as a KE, to be available on the related KER and AOP pages. This planned implementation of the roll-up principle will mitigate content completion issues whereby AOP developers often enter most information on the AOP page itself, including descriptions that pertain to the individual KEs and KERs comprising the AOP. This introduces content sharing issues within the AOP-Wiki because AOP content can be protected whenever Authors choose an All Rights Reserved license. Future expansions to EMOD 3.0 will add new webform-based content entry workflows that shift emphasis to the EMOD data classes, KEs, and KERs that comprise an AOP and add constraints to enforce biological consistency between submissions. In this way, the wider AOP-Wiki user community will still benefit from AOP authors who contribute new insights pertaining to KEs and KERs, while not overburdening those AOP authors with repetitive data entry tasks.

## Beyond EMOD 3.0

While the changes introduced by the EMOD data model offer many advantages, a full expansion of the EMOD 3.0 web application to include the full set of functionalities now in the production AOP-Wiki may take a few years. In the meantime, we are exposing the elements defined in the EMOD model in a LinkML AOP schema hosted under the EHS Data Standards repo[42]. The LinkML schema is undergoing active development. Our hope is that more researchers who use computable approaches to developing AOPs will make use of the EMOD properties when organizing data and evidence in support of new AOPs.

## Figures

### A) Canonical Adverse Outcome Pathway (AOP)

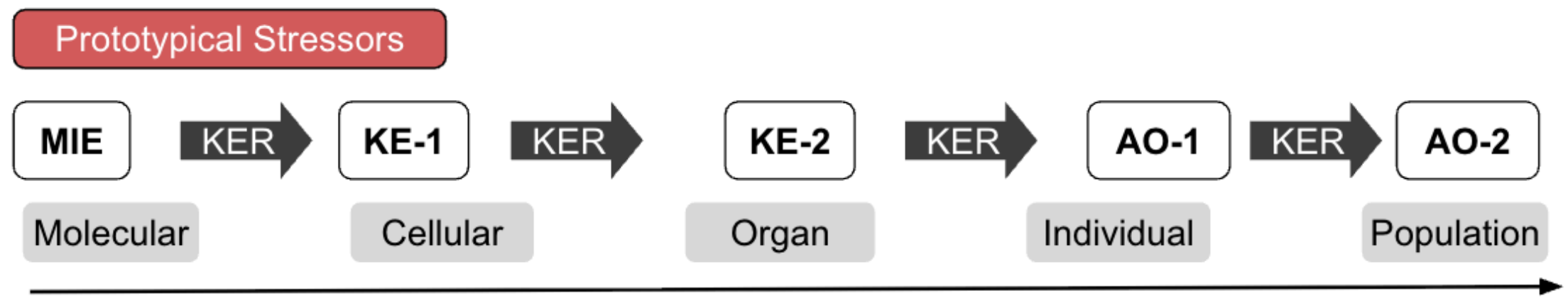

### B) AOP-Wiki EMOD Data Model Classes

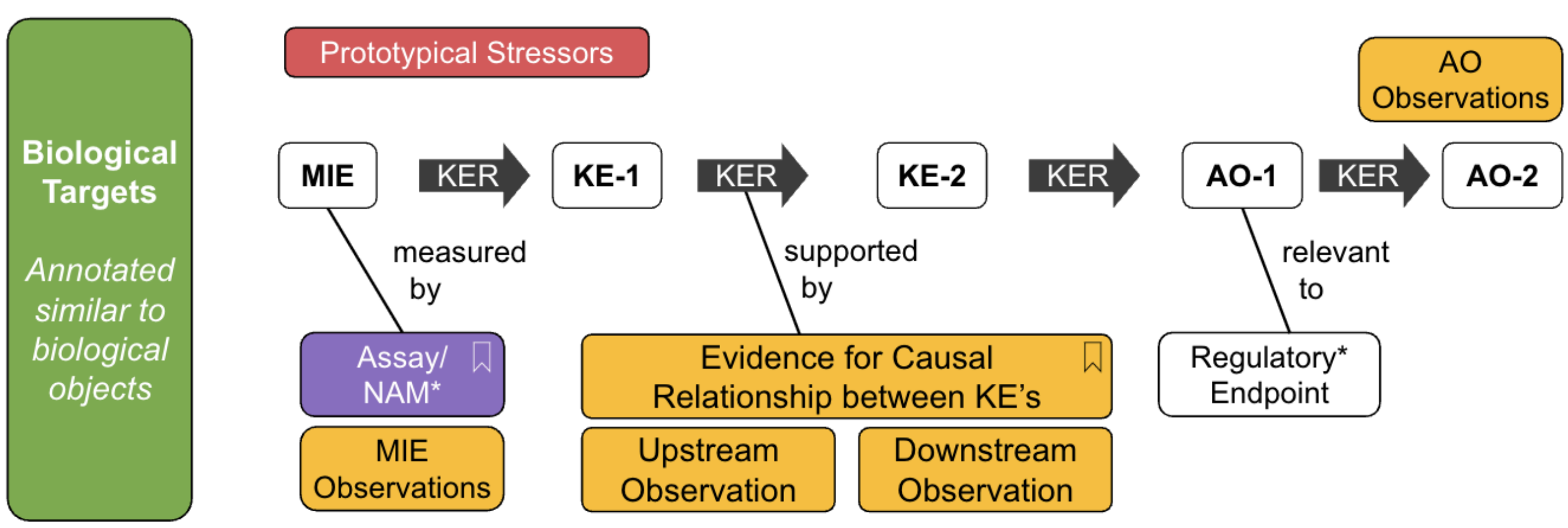

### C) AOP EMOD Observation and Assay Data Class Property Details

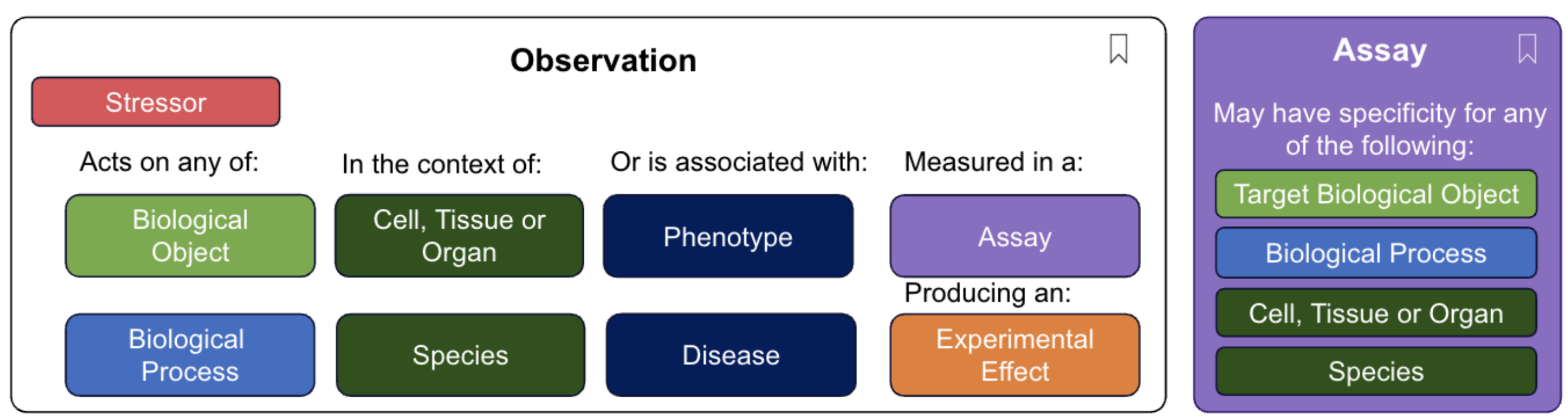

**Figure 1: AOP Framework Properties and AOP-Wiki EMOD Data Model Expansions.** A) Core aspects of the AOP framework. B) AOP-Wiki EMOD data model expansions are color-coded in green, purple and gold. C) Discrete properties on the Observation and Assay data classes enable more granular annotation of AOP-supporting evidence and methods.

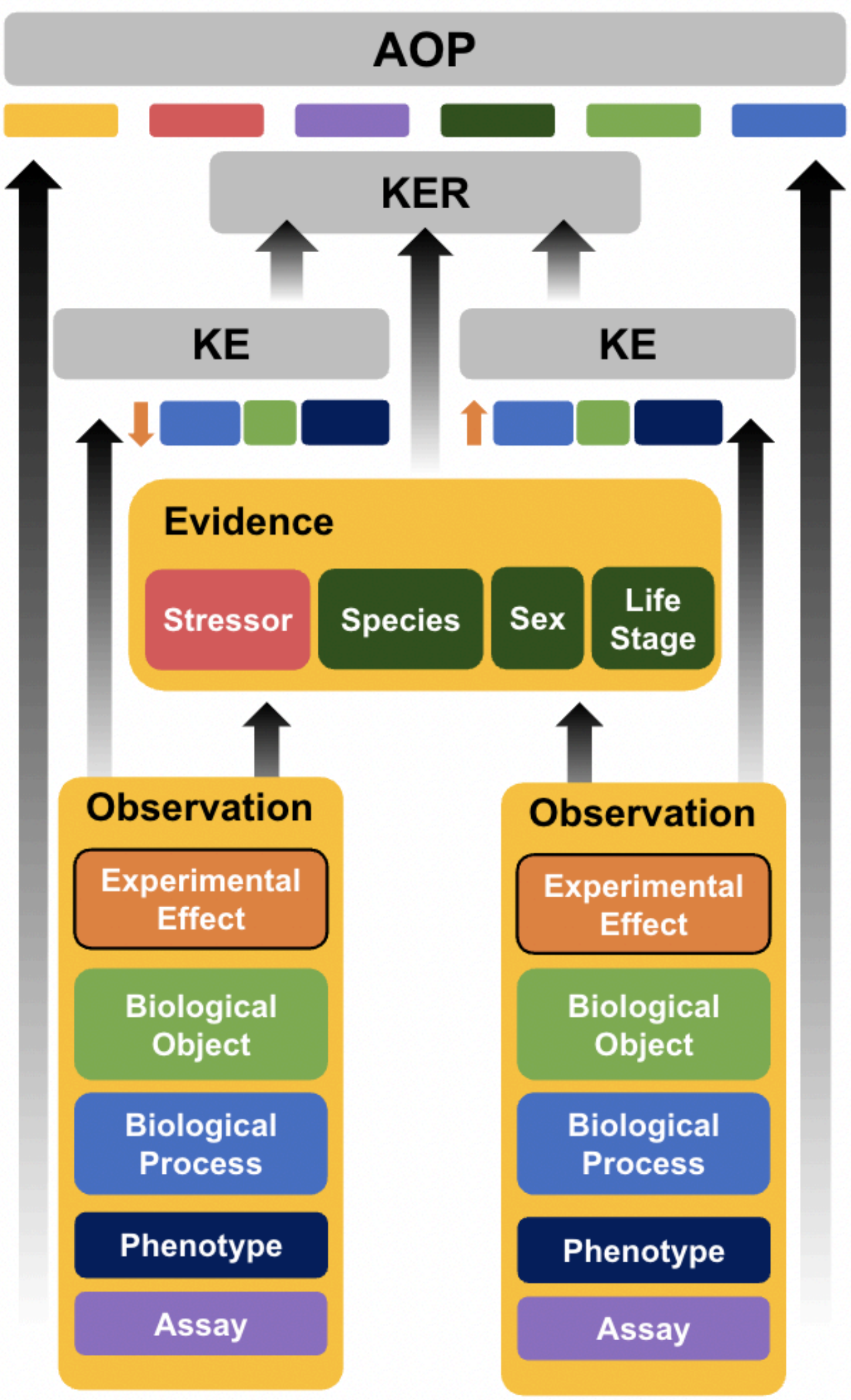

**Figure 2: Visual representation of the AOP EMOD data model principles.** Upward arrows represent the AOP Roll-Up Principle and the boxes for Evidence and Observation represent the properties recommended for each data class. The properties listed on the Evidence and Observation data class boxes are recommendations that align with some KER supporting evidence submissions in the AOP-Wiki. However, more use cases are needed to define an essential set of required properties for each data class.

## Observations

***About AOP Observations:*** *As a data class, AOP Observations are used to organize causal agents, biological elements, and methods of measurement/NAMs in a manner that makes AOP-associated lines of evidence more findable, accessible, and sortable. Observations are linked to Events, which in turn are linked to KERs and AOPs. In the table below:*

- *Stressors with DTXSID's link to individual chemical pages on the US EPA's CompTox Chemicals Dashboard*
- *Bio-ontology terms with available links, link to OBO Ontology records in OntoBee*
- *The bookmark icon at the end of each row serves as a toggle for revealing provenance information for each observation record*

▶ ***Notes on AOP terminology***

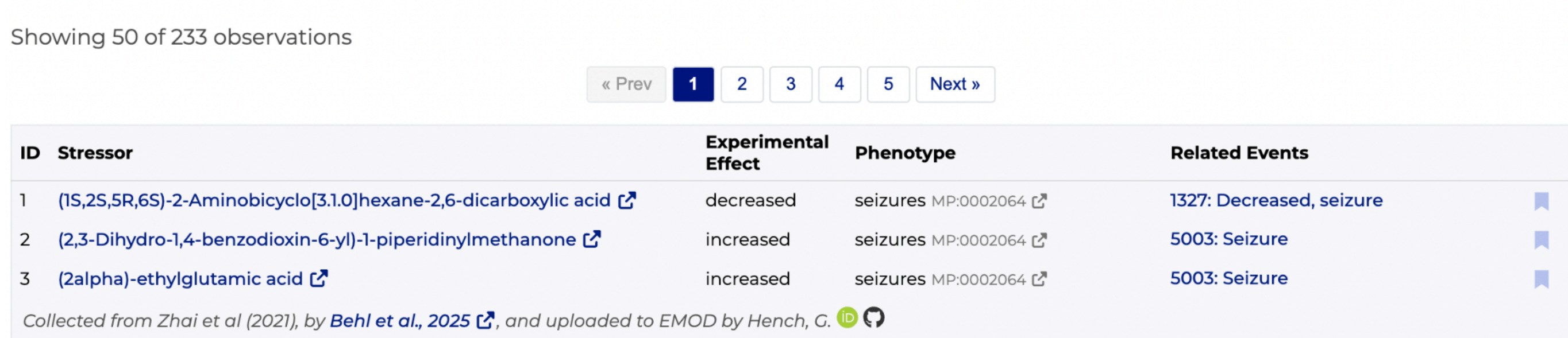

Showing 50 of 233 observations

« Prev 1 2 3 4 5 Next »

| ID | Stressor | Experimental Effect | Phenotype | Related Events |
|---|---|---|---|---|
| 1 | (1S,2S,5R,6S)-2-Aminobicyclo[3.1.0]hexane-2,6-dicarboxylic acid | decreased | seizures MP:0002064 | 1327: Decreased, seizure |
| 2 | (2,3-Dihydro-1,4-benzodioxin-6-yl)-1-piperidinylmethanone | increased | seizures MP:0002064 | 5003: Seizure |
| 3 | (2alpha)-ethylglutamic acid | increased | seizures MP:0002064 | 5003: Seizure |

*Collected from Zhai et al (2021), by **Behl et al., 2025**, and uploaded to EMOD by Hench, G.*

**Figure 3: Observations in EMOD 3.0.** The Observations currently in EMOD 3.0 are based on associations between chemicals and seizure collected by Behl *et al*[30]. The column properties, stressor, experimental effect, and phenotype, are the properties defined on the seizure Observation entries, but the Observation data class allows more properties to be defined.

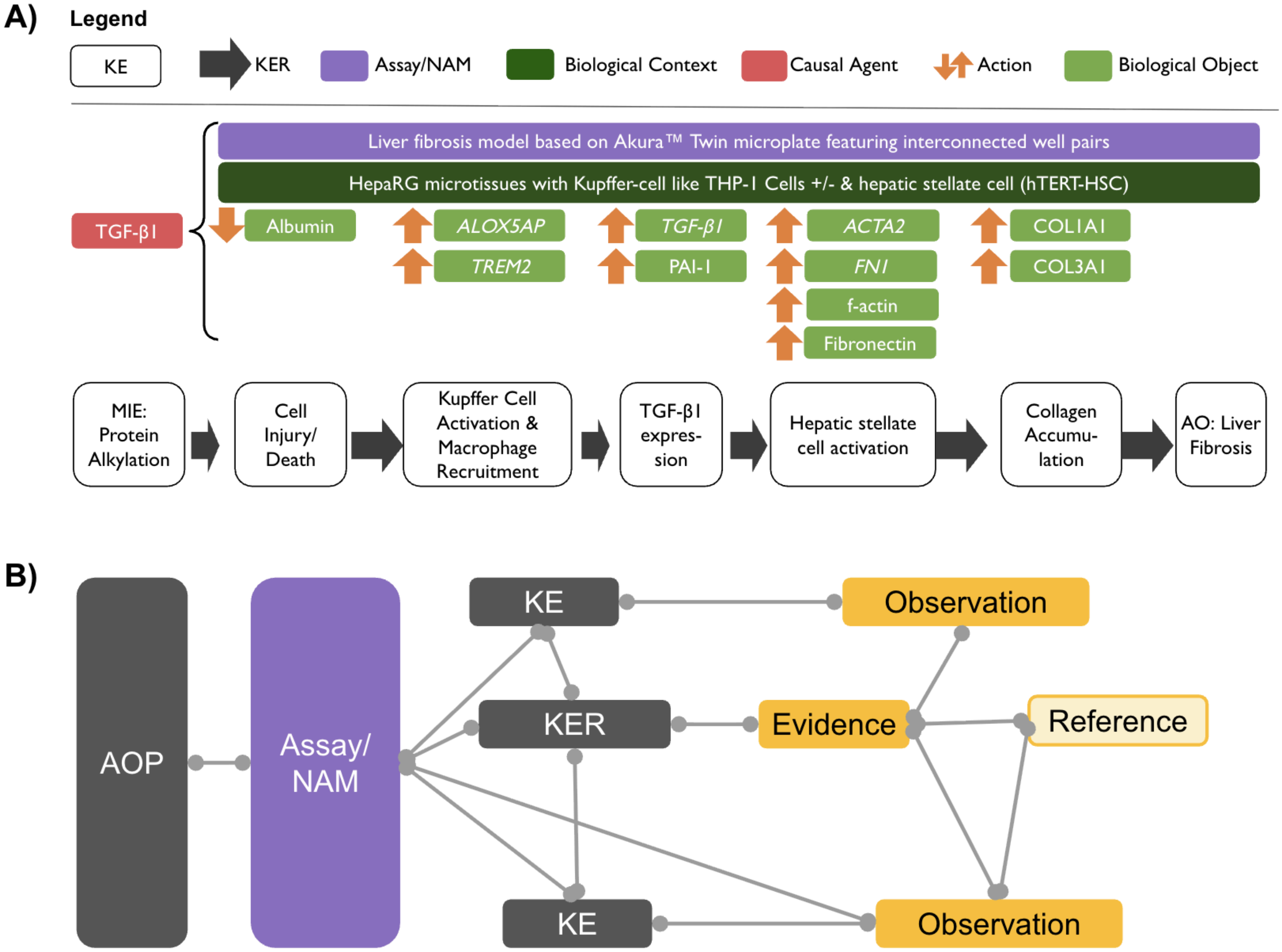

**Figure 4: EMOD Assay/NAM class for Complex *In Vitro* NAMs.** A) Representation of a liver fibrosis microphysiological NAM based on a Liver Fibrosis AOP (AOP #38), with gene expression measurements represented to align with KEs, which would be modelled as Observations. B) Graphical representation of the EMOD data model, showing how Assays can be directly related to AOPs and KERs, as well KEs. Individual endpoint measurements captured within an assay system that maps to a whole AOP or KER are still associated with individual KEs and represented as Observations.

EMOD 3.0 content was copied from the AOP-Wiki v2.8 production site on May 11, 2026.

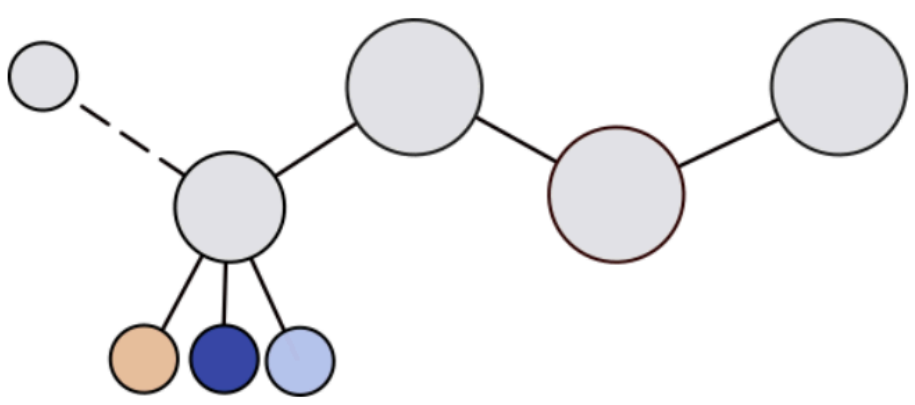

# Welcome to AOP-Wiki EMOD 3.0

*AOP-Wiki EMOD 3.0 is an evidence model prototype app built to demonstrate the value of AOP-Wiki data model expansions and to feature areas where content quality improvement is needed. In this first phase of development, EMOD 3.0 does not include any user login functionality or input forms.*

**CORE ENTITIES**

**Events**
**- LLM-Driven Event Merger Groups**
**- LLM-Driven SODA Event Groups**

**Key Event Relationships (KERs)**

**AOPs**

**USE CASES**

**Seizure Landing Page**
**- Harmonized AOPs**
**- Harmonized Events**
**- Observations**
**- Biological Target Families**

**Lung Fibrosis Event Merger Groups**

**ADVANCED SEARCH**

**Event to KER Search**

**Event Pairs in AOPs**

**KERs with Tabulated Evidence**

**Average Document Completion Scores**

These scores represent the average completion scores for Events, KERs, and AOPs in the database.

| | Events | KERs | AOPs |
|---|---|---|---|
| Without Harmonized Entities | 50.8% | 13.7% | 39.58% |
| With Harmonized Entities | 50.45% | - | 39.33% |

With the first release of AOP-Wiki EMOD 3.0, harmonized AOP and Event entities have been derived from a Seizure Use Case, but not yet populated with new content, which is why the completion scores for the harmonized entities are lower.

**Figure 5: AOP-Wiki EMOD 3.0 Landing Page**

## Events

Filter by name/title/id

Share Results

« Prev 1 2 3 4 5 ... 78 Next »

1942 events (page 1 of 78)

**Select columns to display**

BIOLOGICAL PROPERTIES

☐ Action Term ☐ Cell ☐ Organ ☑ Level of Biological Organization

OTHER METRICS

☑ Has Method Text ☑ % Complete ☐ EC Count ☑ Integration Score

ASSOCIATED AOP PROPERTIES

☑ AOP Count ☐ # AOPs Open for Adoption ☐ # AOPs in OECD AOP Program ☑ # AOPs Endorsed by OECD

- Click the toggle option next to each Event ID to reveal AOPs associated with each Event and the method text.
- AOPs that are Open for Adoption or that are endorsed by the OECD will have **OFA** or **OECD Endorsed** listed at the end of the title, respectively.

| ID ↑ | Name / Title ↕ | Level of Biological Organization | Has Method Text ↕ | % Complete ↕ | Integration Score ↕ | No. Associated AOPs | |
|---|---|---|---|---|---|---|---|
| | | | | | | Total ↕ | OECD Endorsed ↕ |
| 12 or 1,2 | search title… | All | All | >=50 | >=5 | >=1 | >=1 |
| ▶ 3 | Reduction, 17beta-estradiol synthesis by ovarian granulosa cells | Cellular | ✓ | 100% | 17 | 9 | 2 |
| ▶ 8 | Decreased, 3-hydroxyacyl-CoA dehydrogenase type-2 activity | Cellular | | 36.36% | 2 | 1 | 0 |
| ▶ 9 | Activation, 5HT2c | Molecular | | 36.36% | -2 | 1 | 0 |
| ▶ 10 | Acetylcholine accumulation in synapses | Cellular | ✓ | 100% | 12 | 5 | 0 |

**Figure 6: Event Search Page.** A set of default columns display when the Events page first loads, but users are able to modify the columns by choosing from the options under the Select columns to display panel.

**A) Harmonized Events from Seizure Use Case**

1323: Decreased, intracellular serotonin
1398: Increased, intracellular serotonin
→ 5008: Altered Intracellular Serotonin

1325: Decreased, synaptic release
1400: Increased, synaptic release
→ 5009: Altered Synaptic Serotonin Release

1326: Increased, 5-HT3 (5-hydroxytryptamine)
1336: Activation, 5-HT2A (Serotonin 2A)
1401: Decreased, 5-HT3
→ 5010: Activated/Inactivated 5-HT R

**B) Candidate Merger Events From Lung Fibrosis Use Case**

COLLAGEN ACCUMULATION

68: Increase, Collagen accumulation (20)
1275: Collagen Deposition (4)
⟷ *Merge pending — a harmonization strategy has not been recommended*

FIBROBLAST DIFFERENTIATION AND PROLIFERATION

1456: Increase, Differentiation of fibroblasts (2)
1500: Increased, fibroblast proliferation and myofibroblast differentiation (9)
⟷ *Merge pending — a harmonization strategy has not been recommended*

**C) LLM-Driven Candidate Event Merger Groups**

PXR ACTIVATION

239 Activation, Pregnane-X receptor, NR1I2 (11)
245 Activation, PXR/SXR (5)

→ 239: Activation, Pregnane-X receptor, NR1I2
*Recommended Event for harmonization of group*

MITOCHONDRIAL DYSFUNCTION

177 Increase, Mitochondrial dysfunction (50)
178 Mitochondrial electron transport chain, disrupted (5)

**D) LLM-Driven SODA Event Groups**

NMDA RECEPTOR INHIBITION VS OVERACTIVATION

195 Inhibition, NMDARs (11)
388 Overactivation, NMDARs (13)

ESTROGEN RECEPTOR AGONISM ANTAGONISM

111 Agonism, Estrogen receptor (13)
112 Antagonism, Estrogen receptor (10)

**Figure 7: EMOD 3.0 Event Groupings.** A) KEs in the seizure use case were manually curated into groups and a recommended harmonized Event was provided in the process of harmonizing AOPs to support development of NAMs. B) KEs in a lung fibrosis study were combined in groups based on a genomics approach, but a preferred KE wasn't suggested for harmonization, so these groups are treated as Candidate Merger Event groups, rather than harmonized Events. C) An LLM-based approach was used to cluster KEs together based on conceptual similarities conveyed by the Event title. The PXR Activation KE group has a preferred KE that was manually selected. D) An LLM-based approach was used to cluster KEs that fit the SODA KE profiles, meaning same object, different action.

## Supplemental Content

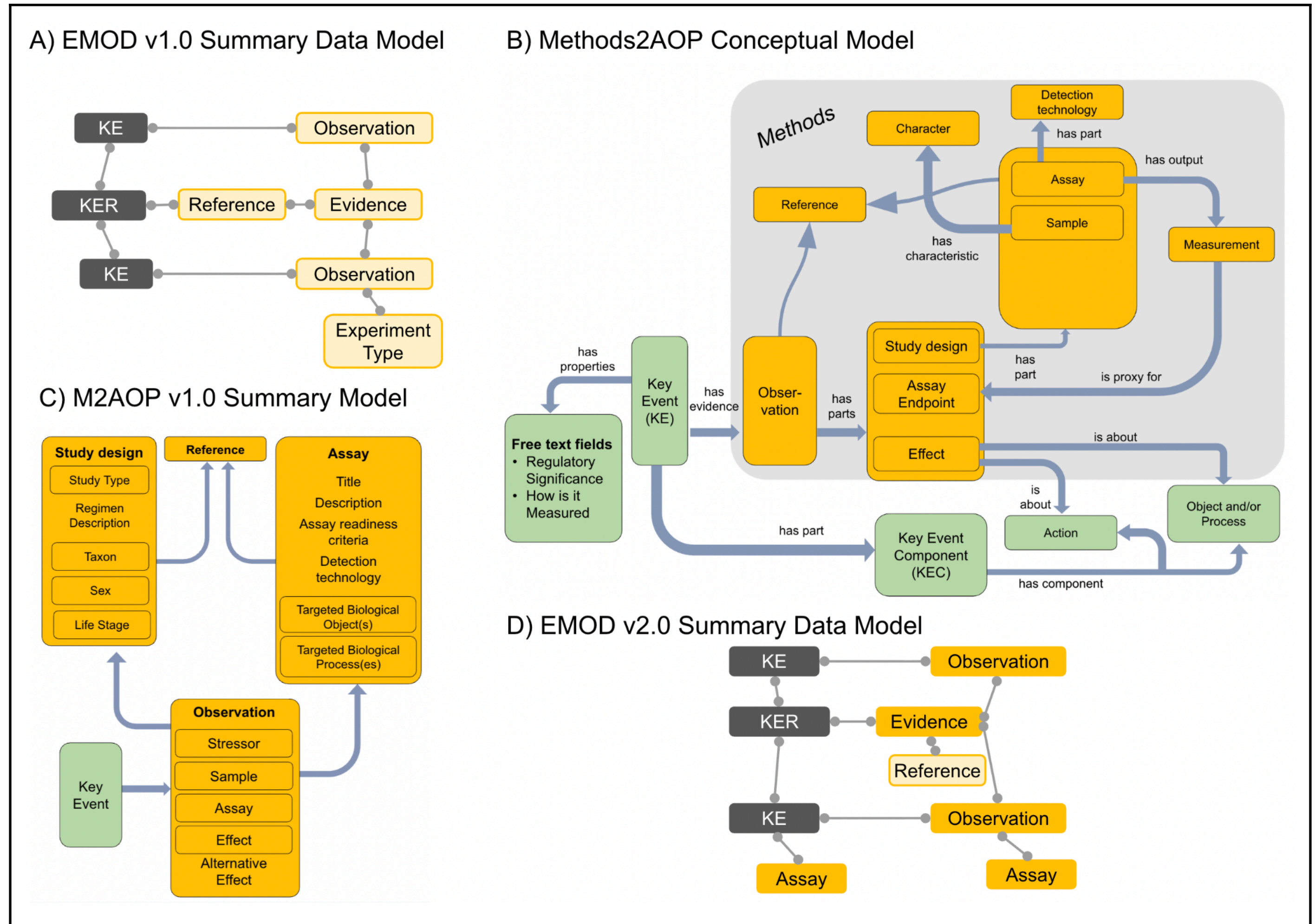

**Supplemental Figure 1: Prototype Data Model Graphics from EMOD 1.0-EMOD 2.0.** A) EMOD v1.0 Summary Model, showing the main data objects introduced to model Evidence for KERs. B) The conceptual model created by members of the Methods2AOP collaboration by summer of 2023. C) Summary model of the M2AOP v1.0 prototype. D) Summary model of the EMOD v2.0 prototype. These figures are based upon work that has previously been shared via presentations in the 2024 workshop on *Roles of the Adverse Outcome Pathway (AOP) framework in incorporating New Approach Methods (NAMs) into the regulatory process*[44] and in a poster presented at the 2024 ISMB meeting[19].

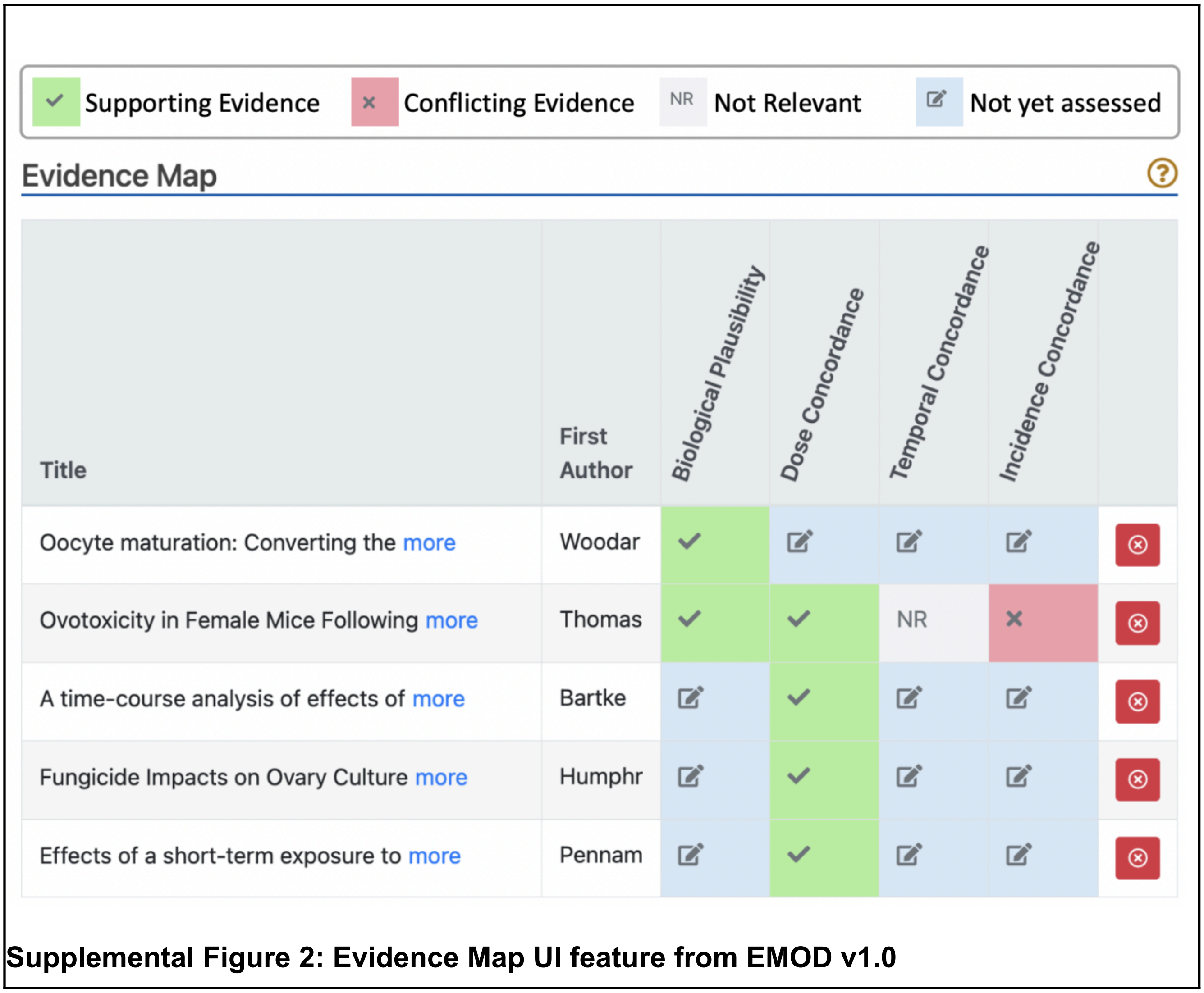

**Supplemental Figure 2: Evidence Map UI feature from EMOD v1.0**

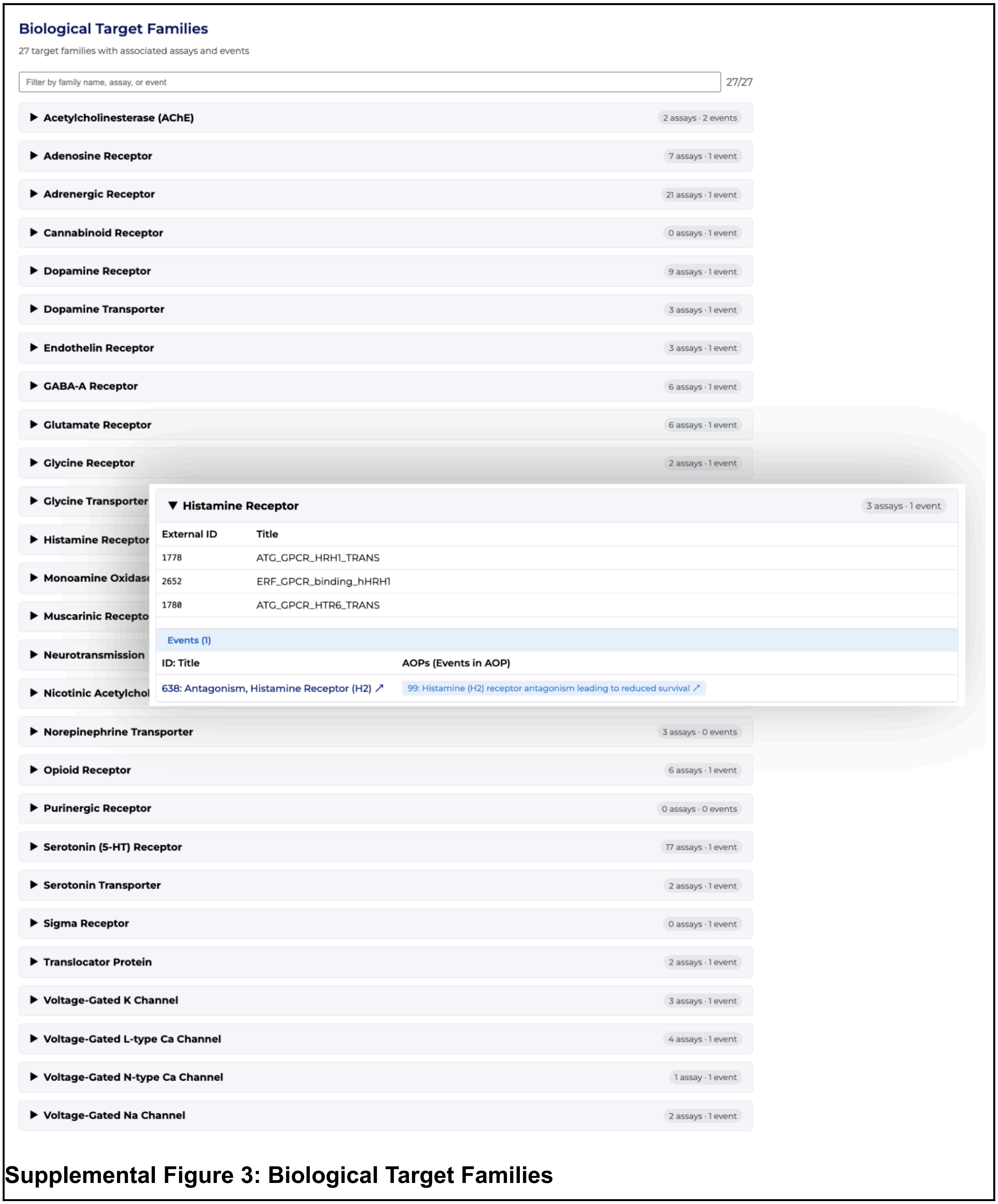

**Supplemental Figure 3: Biological Target Families**